\documentclass{article}

\usepackage{xcolor}

\IfFileExists{neurips_2026.sty}{\def\useneurips{}}{}

\ifdefined\useneurips
  \usepackage[dblblindworkshop,final]{neurips_2026}  
\else
  \usepackage[letterpaper,margin=1.15in]{geometry}
  \usepackage{natbib}
  \usepackage{times}
  \setcitestyle{authoryear,round,citesep={;},aysep={,},yysep={;}}
  
\fi

\usepackage[T1]{fontenc}
\usepackage[utf8]{inputenc}
\usepackage{microtype}
\usepackage{graphicx}
\usepackage{booktabs}
\usepackage{amsmath}
\usepackage{amssymb}
\usepackage{multirow}
\usepackage{array}
\usepackage{placeins}
\usepackage{float}

\makeatletter
\providecommand{\@LN@col}[1]{}
\providecommand{\@LN}[2]{}
\makeatother

\newcommand{\obj}[1]{\ifmmode\text{\texttt{#1}}\else\texttt{#1}\fi}
\newcommand{\splitrepo}{\url{https://github.com/varunagrawal2002/Living-Context-Graph}}
\providecommand{\And}{\\[1ex]}
\providecommand{\AND}{\\[1ex]}

\makeatletter
\@ifpackageloaded{hyperref}{}{%
  \usepackage[colorlinks=true,breaklinks=true]{hyperref}}
\makeatother
\definecolor{darkblue}{rgb}{0,0,.5}
\hypersetup{citecolor=darkblue,linkcolor=darkblue,urlcolor=darkblue,
  bookmarksnumbered=true}
\DeclareRobustCommand{\limref}{\hyperref[sec:limitations]{Limitations}}
\providecommand{\phantomsection}{}

\title{Living Context Graphs:\\ Continual Semantic Learning for\\ Enterprise Data Reasoning}

\providecommand{\workshoptitle}[1]{}
\workshoptitle{Continual Learning in the Era of Foundation Models and Embodied Agents}

\author{%
  Varun Agrawal \\ Genloop \\ \texttt{varun@genloop.ai}
  \And
  Ayush Srivastava \\ Genloop \\ \texttt{ayush.srivastava@genloop.ai}
  \And
  Eshan Gujar \\ Genloop \\ \texttt{eshan@genloop.ai}
  \AND
  Kaushal Kumar \\ Genloop \\ \texttt{kaushal@genloop.ai}
  \And
  Ayush Gupta \\ Genloop \\ \texttt{founder@genloop.ai}
}
   
\begin{document}
\maketitle

\begin{abstract}
Enterprise AI can now perform analytics, reporting, and data-driven workflows,
yet the data foundations it relies on remain hard to operationalize: business
meaning, relationships, metric definitions, and reporting conventions are absent
from database schemas and scattered across undocumented systems and tribal
knowledge. We introduce the Living Context Graph (LCG), a continually learned
context layer that makes existing enterprise data operational for AI without
requiring a copy or transformation of the underlying warehouse. LCG
(\textbf{C1}) bootstraps base semantics and relationships from the data itself,
(\textbf{C2}) builds a concept graph that indexes business context as queries
are served, (\textbf{C3}) writes durable semantic learnings from verified
outcomes, and (\textbf{C4}) withdraws or re-earns context when later outcomes
contradict prior routes. We evaluate LCG through text-to-SQL, a core
data-reasoning task, on a transfer-aware split of three BIRD databases
comprising 499 questions: 353 form the learning history and 146 are held out
and never used to update the context layer. Bootstrapping alone improves
overall accuracy by +4.6 percentage points. Learning only from the history
raises accuracy on the untouched held-out transfer set by +22.6 points (95\% CI
$[+14.4, +30.8]$), exceeding five-shot retrieval from the same history by
+12.3 points. Incremental learning reaches the one-shot endpoint within
observed evaluation noise on every database. Ablating the graph changes
held-out accuracy by at most one question per database but increases per-query
cost 2.3--3.3$\times$ and median latency 1.8--3.4$\times$: continually learned
semantics drive accuracy, while the graph makes that context efficient to
serve.\footnote{The transfer-aware splits and fact-family annotations are
available at
\url{https://github.com/varunagrawal2002/Living-Context-Graph}. The pipeline
itself is not released.}
\end{abstract}

\section{Introduction}
\label{sec:intro}
\paragraph{Capability is not the bottleneck.} AI agents have become increasingly capable of performing analytics, generating reports, and automating workflows. In enterprises, however, capability is often not the limiting factor; the data foundation is. The schema says a column is \texttt{net\_sales\_amt}; it rarely says what the business means by ``net sales,'' which exclusions apply, or how fiscal periods are defined. Those definitions live in dashboards, analyst queries, and tribal knowledge. As a result, ``What were net sales last quarter?'' ought to be easy for a data-reasoning agent, yet an agent with every column and none of this context answers confidently, and wrongly.

\paragraph{Why existing remedies fall short.} To reason reliably over existing data, an agent needs a continuously maintained representation of what the data means, how concepts relate, and which business conventions are valid. Traditional approaches ask an organization to create an ontology, hand-author a semantic layer, transform data into a new representation, or repeatedly supply large amounts of context at inference time. These are costly to deploy over legacy systems and tend to become stale.

\paragraph{The proposal.} We propose the Living Context Graph (LCG), a context layer bootstrapped directly over an existing warehouse that matures through use, with the underlying enterprise data left in place. LCG maintains two complementary forms of context: a semantic store holding human-readable business definitions, relationships, metrics, and corrections; and a concept graph that indexes this context around the vocabulary users actually invoke. At onboarding, the system bootstraps only what can be inferred from the data and schema; during serving, user questions canonicalize recurring concepts and create or strengthen routes into the relevant context; after verified outcomes, the system writes durable semantic learnings, while later failures can withdraw routes or revise prior learnings.

\paragraph{Why text-to-SQL is the vehicle.} Text-to-SQL provides a demanding and externally verifiable test of this idea: it requires an agent to map business language onto schemas, relationships, filters, and reporting conventions, and execution gives a concrete way to measure correctness. We therefore evaluate LCG through a text-to-SQL agent, viewing it as one instance of the broader data-reasoning problem.

\paragraph{The three questions.} The paper asks three questions: how much useful context can be bootstrapped from a warehouse before any user interaction; can verified usage continually add business context that transfers to previously unseen questions, rather than merely memorizing past queries; and can a learned graph serve that context efficiently without becoming the source of accuracy? To answer these, we construct a transfer-aware split across three BIRD databases, learn only from the history split, and evaluate on an untouched held-out set whose required business concepts appear in the history through different questions.

\paragraph{Contributions.} (1) We introduce Living Context Graphs, a no-copy context architecture: a semantic context store is bootstrapped over an existing warehouse and continually revised from verified use. (2) We separate semantic learning from context routing: learned semantics drive transfer gains, while a serving-time concept graph indexes reusable context, reducing cost and latency without materially changing accuracy. (3) We introduce a transfer-aware evaluation protocol over three BIRD databases in which learning and held-out questions share business concepts but not questions. (4) We show that outcome-gated admission is critical: learning from the history raises held-out accuracy from 20/146 to 53/146, while admitting every proposed rule degrades it to 29/146, making validation and forgetting first-class parts of continual semantic learning (\S\ref{sec:results}).

\section{Related work}
\label{sec:related}
\paragraph{Enterprise semantic layers and knowledge representation.}
Industrial semantic layers (e.g., dbt, LookML) encode metric definitions and
relationships but are hand-authored and static. They require upfront modeling and
do not revise themselves from agent outcomes; recent agents mediate NL-to-SQL
through such a layer without learning it~\citep{kim2026semlayer}. Automatic KB
construction~\citep{suchanek2007yago,dong2014kv} and never-ending
learning~\citep{carlson2010nell} pursue automated knowledge representation, making
them relevant to our automatic bootstrapping mechanism (C1), but they typically
pursue this premise without an execution or outcome signal.

\paragraph{Agent memory and continual context.}
Classical continual learning studies parameter
drift~\citep{mccloskey1989catastrophic,kirkpatrick2017ewc,lopezpaz2017gem}, while
a growing line moves adaptation into an external store: agent memory and virtual
context~\citep{park2023generative,packer2023memgpt,xu2025amem,zhou2025memento},
retrieval-based learning~\citep{lewis2020rag,gutierrez2025hipporag2}, and
self-improvement from
trajectories~\citep{madaan2023selfrefine,shinn2023reflexion,wang2023voyager,wang2025awm,zhao2024expel,ouyang2025reasoningbank,suzgun2025cheatsheet,zhang2025ace}.
LCG is positioned as an external, auditable, continually revised context rather
than a weight-learning method. The distinction is not merely that its store is
typed rather than free text or trajectories; it is that LCG learns reusable
business concepts whose effects can be verified on future data-reasoning tasks and
explicitly invalidated. Because it corrects the \emph{representation} rather than
an episode, a fix is reused by every question touching that concept and can go
wrong globally in a way per-episode reflection cannot (\S\ref{sec:e2}, Table~\ref{tab:judge}).

\paragraph{Text-to-SQL, schema linking, and persistent data-agent memory.}
Text-to-SQL has moved from schema-encoding parsers~\citep{wang2020ratsql}
through LLM prompting~\citep{gao2023dailsql} to agentic
pipelines~\citep{wang2024macsql,talaei2024chess,deng2025reforce}, benchmarked on
Spider~\citep{yu2018spider}, BIRD~\citep{li2023bird} and Spider~2.0~\citep{lei2024spider2},
with \emph{schema linking} dominating errors at
scale~\citep{pourreza2023dinsql,wang2024macsql,talaei2024chess,wang2025linkalign}.
Most systems re-solve linking per query and discard
it~\citep{deng2025reforce}; LCG instead writes the resolution back, making linking
a learned, persistent routing problem. This is also the sharpest distinction from
retrieval-augmented prompting~\citep{lewis2020rag}, where placing the right text in
the window is the mechanism: retrieval of the very questions the store learned from
reaches at best 35 of 146 held-out against the learned base's 53 (\S\ref{sec:e1}).
Closest to our work are systems that give a text-to-SQL agent a persistent store:
RubikSQL~\citep{chen2025rubiksql} mines rules by database and SQL profiling,
AgentSM~\citep{biswal2026agentsm} stores execution traces as structured programs,
EvoSQL~\citep{zhou2026evosql} keeps a utility-aggregated candidate memory and
fine-tunes the generator, and others inject tribal knowledge or distil user
feedback into structured
memories~\citep{agarwal2026tribal,cook2025feedback,wang2026l2r}. To our knowledge,
no prior system combines all three of the axes this paper evaluates: a
transfer-aware split, outcome-gated admission and invalidation, and a graph-off
ablation separating semantic content from routing efficiency
(\S\ref{sec:setup}--\ref{sec:results}). We do not offer a direct head-to-head
comparison because these systems are largely closed and use incompatible
evaluation splits, none of which enforces the transfer guarantee ours does.
BIRD~\citep{li2023bird} remains the natural testbed, as its ``external knowledge''
lines name exactly the conventions to acquire, making what must be learned
measurable.

\section{The Living Context Graph}
\label{sec:method}
\begin{figure}[t]
\centering
\includegraphics[width=\textwidth]{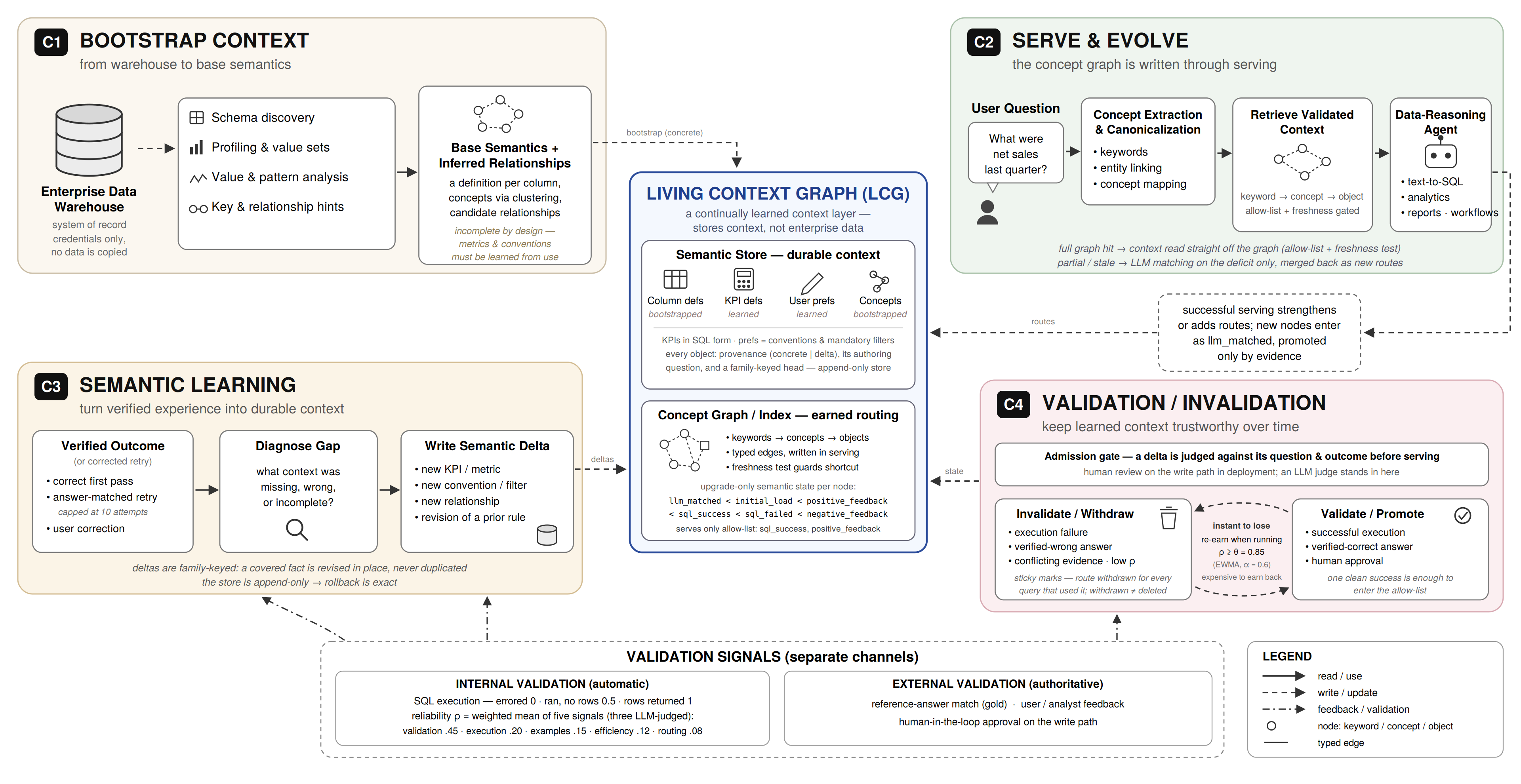}
\caption{
\textbf{The Living Context Graph pipeline.}
\textbf{C1} bootstraps base semantics and inferred relationships from the warehouse, which remains the system of record.
The \emph{Living Context Layer (LCG)} stores semantic context (concepts, rules, metrics, and relationships) as a semantic store and concept graph/index, not a copy of enterprise data.
\textbf{C2} serves questions by extracting and canonicalizing concepts, retrieving validated context, and invoking a data-reasoning agent (e.g., text-to-SQL); successful serving strengthens or adds routes in the graph.
\textbf{C3} converts verified outcomes or corrected retries into durable semantic deltas.
\textbf{C4} validates or invalidates context using both internal signals (execution and reliability) and external signals (reference answers and user/human feedback), allowing context to be re-earned after future validated success.
}
\label{fig:pipeline}
\end{figure}

The LCG is a continually maintained context layer over an existing warehouse. Its semantic store captures business meaning, relationships, metrics, conventions, and corrections in inspectable form; its graph indexes that context around recurring business concepts so an agent retrieves only what a question needs. The two are built and updated together but play different roles: semantic content carries the knowledge, while the graph makes that knowledge efficient to find and serve. The model is held fixed across every condition, so every difference \S\ref{sec:setup} measures is attributable to the context layer. Figure~\ref{fig:pipeline} draws the pipeline; Table~\ref{tab:components} (Appendix~\ref{app:method}) states which part is live in which experiment.

\subsection{C1: the bootstrapped semantic layer}
\label{sec:sem-layer}
Given only warehouse credentials, a dependency-ordered pipeline constructs the initial state: a profiling pass extracts metadata and value sets, an onboarding agent writes one definition per column and maps schema semantics and relationships, and business concepts emerge from clustering these definitions. This produces an incomplete starting context, not an ontology claimed to be correct on day zero. While base semantics can be inferred from the data itself, much cannot be known without use: metric definitions, reporting conventions, organization-specific entity meanings, and business exceptions are strictly missing. Because the initial layer is incomplete and sometimes wrong, the system must learn what was never there and correct what was written badly. Every object carries a provenance tag (\emph{concrete} bootstrap, \emph{delta} learned).

\subsection{C2: graph structure, signal gating and routing}
\label{sec:graph}
The concept graph connects the semantic store to the vocabulary users actually speak, driven entirely by use rather than upfront modeling. Queries introduce keywords and concepts; retrieval canonicalizes them against the semantic context to route the agent. Successful outcomes validate these routes, making them reusable, while failed outcomes withdraw them. Partial misses fall back to semantic retrieval and can create new candidate routes. The core mechanism is earned reusable routing: a node exists because a question needed it, and its influence depends on its verified success. Each node carries a confidence state upgraded only by evidence; the full state order and transition rules are in Appendix~\ref{app:method}.

\subsection{C3: semantic learning} 
\label{sec:learning}
Every verified outcome teaches a reusable business concept or rule: a metric definition, a reporting convention, an entity interpretation, a relationship, a mandatory filter, or a correction to a past mistake. Once a correct answer is known (either on the first pass or through an automated retry loop that explores the data until it matches the expected outcome), the learning agent diagnoses the required context, proposes a durable correction, and writes it back as a \emph{delta}. Rather than memorizing the query, the learner consolidates new information into the existing semantic family. This ensures facts are revised in place rather than duplicated, preserving strict provenance and allowing global reuse across different questions touching the same concept.

\subsection{C4: validation and forgetting} 
\label{sec:validation}
Validation acts as the control loop deciding whether learned context may influence future reasoning. It operates on two axes: internal signals (execution success and reliability checks) and external signals (answer verification or human feedback). Admission is gated: a proposed semantic rule is judged against its question and outcome before it is allowed to serve. Similarly, routing influence is constantly validated. A failed route is withdrawn from active serving but not deleted, preserving its provenance and allowing it to be re-earned upon future validated success.

\section{Experimental setup}
\label{sec:setup}
\paragraph{Fixed model stack.} To isolate the knowledge base rather than
the model, the harness and LLM stack are held fixed:
\obj{gemini-3-flash-preview} at every stage, with identical prompts,
budgets and decoding settings across all conditions.

\paragraph{Datasets.} We run the full protocol on three databases of the
BIRD development set~\citep{li2023bird}: \obj{card\_games},
\obj{toxicology} and \obj{thrombosis\_prediction}, 499 questions in all,
353 learning-history and 146 held-out under the splits of
\S\ref{sec:split}. These are the only databases among the 11 in the dev
set meeting all four protocol requirements
(Table~\ref{tab:screen}, Appendix~\ref{app:discussion}). Each database is
an independent continual-learning problem; pooled numbers micro-average
over the union.

\paragraph{External knowledge.} BIRD ships a database description and a
per-question \obj{evidence} hint. $\ddagger$ conditions receive both, all
others neither; since no enterprise ships a per-question hint, the
hint-free conditions are primary. The learning step never receives either:
it consumes only the question, its answer-matched verdict and the retry
trace (\S\ref{sec:learning}).

\paragraph{Scoring.} BIRD execution accuracy, gold rows against predicted
rows as sets. A longer timeout and MySQL-side execution make numbers
comparable only within this paper (Appendix~\ref{app:method}).

\subsection{A transfer-aware split}
\label{sec:split}
A system that merely memorised its own corrections would look like it had
learned, so the claim needs a split where transfer is the only thing that
can produce a gain. What the mechanism transfers is business context: a
metric definition, reporting rule, entity interpretation, or relationship
learned from one interaction and reused on another. We therefore group
questions by normalised semantic concept families (\emph{fact families}
in the released artifacts) rather than shared columns or lexical
similarity, and a held-out question is a transfer target only when every
business fact it requires has a teacher in the learning history while the
question itself remains unseen. The split tests reachable transfer, not
general BIRD accuracy.

For \obj{card\_games} we annotated all 191 questions with the
phrase-level facts their gold answers depend on, normalised to keys so
the same concept collides across questions: 60 families, 39 of them
\emph{transfer families} shared between a held-out target and at least
one teacher, 21 private to single learning-history questions. The split
is then arithmetic on those keys. The \textbf{held-out transfer split
(57)} contains transfer targets only: every needed fact (one or two per
question) has a teacher, no held-out question carries a private fact, and
ties within a family were broken at random. The \textbf{learning history
split (134)} plays the role of a deployment's query history: 70 teachers
and 64 \emph{independents} whose facts no held-out question needs,
serving as regression probes. Coverage is verified after placement; the
single-pass author annotation is the only manual step.

The same procedure yields \textbf{101/44} on \obj{toxicology} (33
families, 18 transfer; 59 teachers, 42 independents) and \textbf{118/45}
on \obj{thrombosis\_prediction} (142 families, 71 transfer; 102 teachers,
16 independents; Table~\ref{tab:independents}; worked examples,
Appendix~\ref{app:discussion}). Annotations and splits are released at
\splitrepo.

\paragraph{Metrics.} Beyond execution accuracy: per-query cost and
latency; \emph{context recall}, the fraction of the gold query's columns
placed in the prompt; and graph \emph{full-} and \emph{partial-hit} rates per Eq.~\eqref{eq:resolved} (Appendix~\ref{app:method}).

\section{Experiments}
\label{sec:exp}
\label{sec:results}

Three experiments address the questions posed in \S\ref{sec:intro}. Exp~1
establishes what context is obtainable without usage, and bounds what supplying
the learning history verbatim achieves. Exp~2 measures whether verified use
yields context that transfers to unseen questions, at what cost in interference,
and what the admission gate contributes. Exp~3 measures whether the same
knowledge delivered incrementally reaches the one-shot endpoint, and what graph
routing contributes to the cost of serving it.

\paragraph{Interpretation of the reported quantities.}
Held-out questions were selected to require conventions absent from the schema
(\S\ref{sec:split}); held-out accuracy is therefore a floor by construction and
is not a general statement about accuracy on BIRD.

\subsection{Exp~1: bootstrapped context and retrieval baselines}
\label{sec:e1}
\paragraph{Conditions.}
Five static, graph-free conditions differing only in the context supplied.
\emph{(a)~Raw schema}: names, columns and types, with the semantic tools
withheld. \emph{(b)~Bootstrapped catalog}: the semantic layer generated per \S\ref{sec:sem-layer}, which contains no KPIs and no preferences.
\emph{(r)} and \emph{(s)}: few-shot baselines supplying the five most similar
learning-history questions with their gold SQL, over the raw schema (r) or over
the catalog (s); retrieval is a BM25-and-dense hybrid fused by reciprocal rank,
with nothing from the held-out split indexed (Appendix~\ref{app:method}).

\begin{table}[t]
\centering
\caption{\textbf{Accuracy across all conditions.} Rows (a)--(d) involve no
learning, so the \emph{all} column is the appropriate reading there; in rows
(e)--(h) the learning-history questions contributed their own rules, so
\emph{held-out} measures transfer. Rows (r) and (s) ($k{=}5$; sweep over $k$ in
Table~\ref{tab:ksweep}, Appendix~\ref{app:method}) are evaluated on held-out
only, since a learning-history question would retrieve itself. Oracle rows learn
from every question, including held-out, and are leaked ceilings rather than
results. $^{\ddagger}$ marks conditions supplied with \textbf{external
knowledge}, BIRD's per-question \obj{evidence} line together with the database
description.}
\label{tab:base}\label{tab:learned}
\footnotesize
\setlength{\tabcolsep}{2pt}
\begin{tabular}{@{}l l rrr rrr rrr rrr@{}}
\toprule
& & \multicolumn{3}{c}{\textbf{\obj{card\_games}}} & \multicolumn{3}{c}{\textbf{\obj{toxicology}}} & \multicolumn{3}{c}{\textbf{\obj{thrombosis\_pred.}}} & \multicolumn{3}{c}{\textbf{pooled}} \\
\cmidrule(lr){3-5}\cmidrule(lr){6-8}\cmidrule(lr){9-11}\cmidrule(l){12-14}
& \textbf{Knowledge-base state} & {\scriptsize hist.} & {\scriptsize h.o.} & {\scriptsize all} & {\scriptsize hist.} & {\scriptsize h.o.} & {\scriptsize all} & {\scriptsize hist.} & {\scriptsize h.o.} & {\scriptsize all} & {\scriptsize hist.} & {\scriptsize h.o.} & {\scriptsize all} \\
& & {\scriptsize 134} & {\scriptsize 57} & {\scriptsize 191} & {\scriptsize 101} & {\scriptsize 44} & {\scriptsize 145} & {\scriptsize 118} & {\scriptsize 45} & {\scriptsize 163} & {\scriptsize 353} & {\scriptsize 146} & {\scriptsize 499} \\
\midrule
\multicolumn{14}{@{}l}{\textbf{No learning}} \\
(a) & Raw schema only & 61 & 5 & \textbf{66} & 69 & 5 & \textbf{74} & 21 & 9 & \textbf{30} & 151 & 19 & \textbf{170 (34.1\%)} \\
(r) & \ \ $+$ 5 retrieved examples & -- & 13 & -- & -- & 8 & -- & -- & 14 & -- & -- & 35 & -- \\
(b) & Bootstrapped catalog & 71 & 6 & \textbf{77} & 79 & 4 & \textbf{83} & 23 & 10 & \textbf{33} & 173 & 20 & \textbf{193 (38.7\%)} \\
(s) & \ \ $+$ 5 retrieved examples & -- & 12 & -- & -- & 7 & -- & -- & 14 & -- & -- & 33 & -- \\
(c) & Raw $+$ external knowledge$^{\ddagger}$ & 78 & 13 & \textbf{91} & 74 & 8 & \textbf{82} & 44 & 17 & \textbf{61} & 196 & 38 & \textbf{234 (46.9\%)} \\
(d) & Bootstrapped catalog $+$ external knowledge$^{\ddagger}$ & 83 & 14 & \textbf{97} & 73 & 6 & \textbf{79} & 53 & 17 & \textbf{70} & 209 & 37 & \textbf{246 (49.3\%)} \\
\addlinespace[0.45em]
\multicolumn{14}{@{}l}{\textbf{Learned from the learning history only} : \emph{the system as deployed}} \\
(e) & \ \ $+$ agent-written knowledge base & 105 & \textbf{22} & 127 & 87 & \textbf{12} & 99 & 77 & \textbf{19} & 96 & 269 & \textbf{53} & \textbf{322 (64.5\%)} \\
(f) & \ \ \ \ $+$ external knowledge$^{\ddagger}$ & 104 & \textbf{29} & 133 & 90 & \textbf{13} & 103 & 79 & \textbf{18} & 97 & 273 & \textbf{60} & \textbf{333 (66.7\%)} \\
\addlinespace[0.45em]
\multicolumn{14}{@{}l}{\textbf{Learned from all questions} : \emph{upper bound: held-out leaked}} \\
(g) & \ \ \emph{Oracle} & 101 & 38 & 139 & 86 & 22 & 108 & 74 & 27 & 101 & 261 & 87 & 348 (69.7\%) \\
(h) & \ \ \ \ $+$ external knowledge$^{\ddagger}$ & 104 & 37 & 141 & 91 & 25 & 116 & 86 & 30 & 116 & 281 & 92 & 373 (74.7\%) \\
\bottomrule
\end{tabular}
\end{table}

\paragraph{Effect of the bootstrapped semantic layer.}
Bootstrapping the semantics per \S\ref{sec:sem-layer} raises all-question
accuracy from 34.1\% to 38.7\% (170 to 193 of 499, Table~\ref{tab:base}). On
held-out questions the effect is negligible, 13.0\% to 13.7\% (19 to 20 of 146).
The asymmetry is a property of the split rather than of the layer: held-out
questions were selected to depend on conventions the schema does not carry, and
no amount of schema exploration recovers them. The bootstrap supplies structure,
which \S\ref{sec:sem-layer} states is the limit of what warehouse profiling can
establish.

\paragraph{Retrieval baselines.}
Retrieving the five most similar learning-history questions raises held-out accuracy from 13.0\% to 24.0\% (19 to 35 of 146) over the raw schema, validating the split construction of \S\ref{sec:split}. Over the bootstrapped catalog, retrieval reaches 22.6\%. Crucially, \emph{learning} from this supervision rather than supplying it at inference achieves 36.3\% (condition (e)), a $+12.3$ point gain over the strongest retrieval baseline. This gap remains consistent across databases and persists across sweeps of $k \in \{3,5,10\}$ (Table~\ref{tab:ksweep}). Note that rows (r) and (s) act as upper bounds, as their BM25 variant was tuned on held-out labels, an advantage the semantic layer lacked.

\subsection{Exp~2: transfer from verified outcomes}
\label{sec:e2}
\paragraph{Procedure.}
Condition (e) of Table~\ref{tab:learned}: the learning step consumes the
learning-history questions and their verified outcomes and writes back metric
definitions and reporting conventions; on \obj{card\_games} every learned object
is one of these two types (Table~\ref{tab:sem-objects}). The graph remains
disabled, so any difference is attributable to semantic content.

\paragraph{Transfer to held-out questions.}
The primary contrast is hint-free, (b) against (e), corresponding to the deployed
setting of \S\ref{sec:setup}: held-out accuracy rises from 13.7\% to 36.3\% (20
to 53 of 146), a gain of $+22.6$ points, with a 95\% CI of $[+21,+45]$ questions
under a paired bootstrap over the 146 held-out questions. The effect is present
on each database: 10.5\% to 38.6\% on \obj{card\_games}, 9.1\% to 27.3\% on
\obj{toxicology}, and 22.2\% to 42.2\% on \obj{thrombosis\_prediction}
($+28.1$, $+18.2$ and $+20.0$ points; the smallest gain is $+8$ of 44). The harness and model are held fixed across conditions, so the gain is attributable to the context layer. We report
absolute deltas rather than multipliers, since a ratio computed on this floor is
unstable.

\paragraph{Character of the learned objects.}
The store is legible, and its contents indicate that what transfers is business
convention rather than restated schema. On \obj{card\_games} at $t{=}20$ it holds
seven distinct metric definitions and 44 distinct preferences written from 20
questions: a metric encoding a specific ratio over two boolean columns; a preference
fixing the exact comma-separated form in which \obj{availability} stores multiple
platforms; a rule recording that a ``set number'' resolves against
\obj{set\_translations.id}. Each object carries its authoring question, so every
answer is traceable to the rules that produced it and every rule to the query
that taught it.

\paragraph{Cost of the supervision.}
Questions answered incorrectly on the first pass enter the answer-matched retry
loop of \S\ref{sec:learning}, capped at ten attempts. Between 80.6\% and 86.4\%
of eligible questions are corrected, at a mean of 2.9 to 4.1 attempts and never
more than eight; questions not corrected within the cap are not learned from
(Table~\ref{tab:retry}, Appendix~\ref{app:results}). This verdict is the only
supervision the loop consumes and is stronger than the signal a deployment
supplies (\limref).

\paragraph{Interference from marginal rules.}
Learning additionally from the 146 held-out questions improves the questions it
targets, but reduces accuracy on the learning history on every database: 101
against 105, 86 against 87, and 74 against 77. In continual-learning terms this
is negative backward transfer~\citep{lopezpaz2017gem}, $-8/353$ pooled or $-2.3$
points at matched supervision (Appendix~\ref{app:discussion}).

\paragraph{Contribution of the admission gate.}
Re-running condition (e) with the admission judge disabled, so that every
proposed learning serves, reduces pooled held-out accuracy from 36.3\% to 19.9\%
(53 to 29 of 146) and learning-history accuracy from 269 to 207 of 353, while the
store grows from 410 to 488 KPI definitions and from 354 to 421 preferences
(Table~\ref{tab:judge}, Appendix~\ref{app:results}). The direction is consistent
across all three databases: the unfiltered store holds more rules and performs
worse, so admission rather than volume carries the gain. 

\subsection{Exp~3: incremental delivery and the cost of serving}
\label{sec:e3}
\paragraph{Protocol.}
Each run begins from the bootstrapped catalog with no learned rules and an empty
graph. Learning-history questions arrive in random batches of approximately 20
under a fixed seed, stratified over semantic concept families so that no batch
concentrates teachers. Each batch is answered live; correct answers are learned
as they stand and incorrect ones retried until the returned answer matches, then
learned from (full protocol in Appendix~\ref{app:results}). No \obj{evidence}
line or database description is supplied. The state is then frozen and the
held-out questions evaluated against it, producing no graph or semantic write.

\begin{table}[t]
\centering
\caption{Exp~3 endpoints: held-out questions evaluated against frozen state at
the start and end of each stream (median latency). $^{\dagger}$: the final store,
graph disabled; $^{\ast}$: the stream continued through its own held-out
questions, so the row is a leaked upper bound. Every checkpoint, split by serving
path, is Table~\ref{tab:e3full} (Appendix~\ref{app:results}).}
\label{tab:e3}
\small
\setlength{\tabcolsep}{5pt}
\begin{tabular}{@{}rrrrrr@{}}
\toprule
$t$ & held-out acc. & \$/query & latency (s) & ctx. recall & full hit \\
\midrule
\multicolumn{6}{@{}l}{\obj{card\_games}: 57 held-out, 134 learning-history questions} \\
0 & \ 7 \ (12.3\%) & 0.0166 & 165.1 & 87.0\% & \ 0.0\% \\
134 & 22 \ (38.6\%) & 0.0448 & 110.0 & 90.2\% & 78.9\% \\
134$^{\dagger}$ & 23 \ (40.4\%) & 0.1477 & 373 & 95.6\% & -- \\
191$^{\ast}$ & 36 \ (63.2\%) & 0.0402 & 182.8 & 97.0\% & 91.2\% \\
\midrule
\multicolumn{6}{@{}l}{\obj{toxicology}: 44 held-out, 101 learning-history questions} \\
0 & \ 5 \ (11.4\%) & 0.0137 & 172 & 88.1\% & \ 0.0\% \\
101 & 12 \ (27.3\%) & 0.0467 & 181 & 93.7\% & 63.6\% \\
101$^{\dagger}$ & 11 \ (25.0\%) & 0.1084 & 326 & 98.9\% & -- \\
145$^{\ast}$ & 22 \ (50.0\%) & 0.0234 & \ 98 & 94.1\% & 93.2\% \\
\midrule
\multicolumn{6}{@{}l}{\obj{thrombosis\_prediction}: 45 held-out, 118 learning-history questions} \\
0 & \ 8 \ (17.8\%) & 0.0174 & 212 & 92.7\% & \ 0.0\% \\
118 & 18 \ (40.0\%) & 0.1025 & 277 & 99.4\% & 73.3\% \\
118$^{\dagger}$ & 19 \ (42.2\%) & 0.2755 & 539 & 95.6\% & -- \\
163$^{\ast}$ & 28 \ (62.2\%) & 0.0385 & 126 & 99.4\% & 97.8\% \\
\bottomrule
\end{tabular}
\end{table}

\paragraph{Equivalence of incremental and one-shot delivery.}
\obj{card\_games} rises from 12.3\% (7 of 57) at $t{=}0$ to 38.6\% (22 of 57) at
$t{=}134$, matching its one-shot score exactly (Table~\ref{tab:e3};
Figure~\ref{fig:curve}). \obj{toxicology} likewise ends at its one-shot score,
27.3\% (12 of 44). \obj{thrombosis\_prediction} ends at 18/45 against a one-shot
19/45. Knowledge delivered incrementally is therefore worth as much as knowledge installed at once, subject to the same learning and serving cost; the leaked continuations in Table~\ref{tab:e3} are upper bounds only (Appendix~\ref{app:results}).

\paragraph{Checkpoint variability.}
The per-checkpoint curves are not monotone, but every decrease is one to two
questions, within single-evaluation noise (\limref), so the
shape is reported as observed and carries no claim; provenance nonetheless
attributes it, the decrease at $t{=}81$ on \obj{toxicology} resolving to two
admitted rules and recovering by $t{=}101$. The robust feature is the plateau,
which occurs once the learning history ceases to teach facts a held-out question
requires (Limitations). Invalidation is visible in coverage rather than accuracy:
a failing batch withdraws nodes, and the affected questions re-route through the
semantic engine until those routes are re-earned
(Figure~\ref{fig:hitpaths}). Of break events with an opportunity to recover, 61\% do
(Table~\ref{tab:recovery}, Appendix~\ref{app:results}), an upper bound since a
verdict flip registers as a break and a recovery alike.

\paragraph{Contribution of graph routing.}
Re-evaluating each stream's final store with the graph disabled leaves accuracy
substantially unchanged, at 23/57, 11/44 and 19/45 against 22/57, 12/44 and 18/45
with routing enabled (Table~\ref{tab:e3}). Cost separates the
two conditions more precisely: the same questions cost \$0.1477,
\$0.1084 and \$0.2755 per query without the graph against \$0.0448, \$0.0467 and
\$0.1025 with it ($2.3$--$3.3\times$), at median latencies of 373, 326 and 539\,s
against 110, 181 and 277\,s ($1.8$--$3.4\times$). The mechanism is that of
\S\ref{sec:graph}: a full hit bypasses the matching pass, and by each stream's
end 63.6--78.9\% of held-out questions are full hits, 97.7--100\% at least
partial. Full-hit context recall rises from 58--82\% early in each stream to 92--99\% at
its end, converging with and in places exceeding that of the fallback matching pass (Figure~\ref{fig:hitpaths}). 

\FloatBarrier
\phantomsection
\label{sec:limitations}
\section*{Limitations}
Five limitations bound what these experiments show. \emph{Data volume and coverage}: Evaluated on 499 questions across three databases, held-out sets are specifically designed to test knowledge transfer rather than reflect natural question distributions, meaning success requires the learning history to contain the needed facts (\S\ref{sec:split}). \emph{History saturation}: Accuracy plateaus early ($t{=}40$ on \obj{toxicology}, $t{=}100$ on \obj{card\_games}) because later questions repeat concepts, so this plateau reflects the dataset's limits, not the mechanism's maximum capacity. \emph{Evaluation variance}: Due to LLM costs, each condition uses a single run, but while run-to-run noise affects 1--2 questions per database, our primary pooled findings (e.g., $+33$ questions, 95\% CI $[+21,+45]$) are an order of magnitude larger than this variance. \emph{Supervision strength}: Learning relies on verified correct SQL via answer matching, providing stronger supervision than the passive ``no-complaint'' feedback typical of real-world deployments. \emph{Generator-judge correlation}: The admission gate is crucial (Table~\ref{tab:judge}, Appendix~\ref{app:results}) and relies on an LLM judge; while filtering drives our results, we have not measured the judge's agreement with human reviewers.

\paragraph{Broader impact.} Living Context Graphs (LCG) enable organizations to directly operationalize enterprise data for AI, bypassing the months traditionally required for upfront ontology design or large data migrations. While explicitly extracting tribal knowledge from analysts' query patterns (Appendix~\ref{app:method}) is highly valuable, this automation fundamentally risks displacing the very analysts required to sanity-check the system's contents. However, LCG still requires human review on the write path (\S\ref{sec:validation}) to ensure that ongoing learnings remain in sync with the enterprise. Coupled with the fact that every learned object carries the query that taught it (Table~\ref{tab:sem-objects}), the resulting knowledge store remains highly legible, provenance-tagged, and fully auditable by domain experts.

\section{Conclusion}
We presented Living Context Graphs (LCG), a continually learned context layer for making existing enterprise data operational for AI. Rather than requiring an organization to fully model its data upfront, LCG bootstraps the context that can be inferred from a warehouse, then adds and revises business meaning through verified use. In our text-to-SQL evaluation, learning from a 353-question history raises held-out accuracy by 22.6 percentage points, while incremental learning reaches the same endpoint as one-shot learning within observed evaluation noise. Ablating the graph increases serving cost by $2.3$--$3.3\times$ and latency by $1.8$--$3.4\times$. This separation is central to the design: semantics capture organizational knowledge; the graph makes it efficient to operationalize. The remaining challenge is deciding which learnings deserve to become durable organizational knowledge; evaluating noisier enterprise feedback and broader data-reasoning agents is the next step.

\begin{ack}
This work was funded by Genloop. The authors declare no competing interests.
\end{ack}

\newpage

\appendix

\section{Method detail}
\label{app:method}

\begin{table}[H]
\centering
\caption{Which component is live in which experiment; model stack, prompts,
budgets and SQL agent are identical in every cell. C2 and C4's invalidation loop are disabled in
Exps~1 and~2, so nothing those experiments show can be credited to routing; C4's
admission gate is live wherever C3 writes, which is what Table~\ref{tab:judge}
ablates.
External knowledge appears only in the conditions marked $\ddagger$, never
during a stream.}
\label{tab:components}
\small
\begin{tabular}{@{}llccc@{}}
\toprule
& & \textbf{Exp~1} & \textbf{Exp~2} & \textbf{Exp~3} \\
& & {\footnotesize\emph{bootstrap}} & {\footnotesize\emph{one-shot learning}} & {\footnotesize\emph{online stream}} \\
\midrule
\textbf{C1} & bootstrapped semantic layer (\S\ref{sec:sem-layer}) & \checkmark & \checkmark & \checkmark \\
\textbf{C2} & signal-gated graph routing (\S\ref{sec:graph}) & \textendash & \textendash & \checkmark \\
\textbf{C3} & semantic learning loop (\S\ref{sec:learning}) & \textendash & \checkmark & \checkmark \\
\textbf{C4a} & admission gate (\S\ref{sec:validation}) & \textendash & \checkmark & \checkmark \\
\textbf{C4b} & graph invalidation loop (\S\ref{sec:validation}) & \textendash & \textendash & \checkmark \\
\addlinespace[0.35em]
\multicolumn{2}{@{}l}{\emph{learning is offline over the whole train set at once}} & \textendash & \checkmark & \textendash \\
\multicolumn{2}{@{}l}{\emph{learning is incremental, in batches of 20}} & \textendash & \textendash & \checkmark \\
\bottomrule
\end{tabular}
\end{table}

\begin{table}[H]
\centering
\caption{\textbf{The four object types.} Types and fields are
identical on every database; counts are \obj{card\_games}', the running example.
Structure is derivable from the warehouse and generated once; business meaning is
not, and is acquired continually from use. A provenance field on every object
separates the 138 bootstrapped objects from the 178 the learning step wrote and
names the question that taught each one, which is what makes every number in
\S\ref{sec:exp} attributable to a rule and a query.}
\label{tab:sem-objects}
\small
\begin{tabular}{@{}p{0.20\textwidth}p{0.74\textwidth}@{}}
\toprule
\textbf{Object} & \textbf{What its fields carry} \\
\midrule
\obj{Column defs} \newline \emph{bootstrapped} (115) &
Business meaning written toward use (how the column bears on computing KPIs), a
longer definition, prose on how a value is composed, sample values, and how a
null must be read here. Two fields are conditional: closed-domain value lists
(26/115) and a generated converter for awkward date encodings (2/115) \\
\addlinespace[0.35em]
\obj{KPI defs} \newline \textbf{never} \textbf{bootstrapped} (95) &
The metric expression in SQL form, so it is applied verbatim rather than
paraphrased; the fully qualified columns it references; and an origin marker that
reads \obj{created-with-sql} on all 95, every KPI came from a query that
executed, none was authored \\
\addlinespace[0.35em]
\obj{User prefs} \newline \textbf{never} \textbf{bootstrapped} (83) &
The rule text (a convention or a mandatory filter the SQL must respect), its
scope - a real table for table-scoped rules, an org-wide marker otherwise and the concepts the rule asked to attach to \\
\addlinespace[0.35em]
\obj{Concepts} \newline \emph{bootstrapped} (23) &
Three membership lists (columns, KPIs, preferences) that make a concept the unit
of retrieval. The last is the list routing walks, so a preference is reachable
only through a concept it attached to \\
\addlinespace[0.35em]
\emph{on every object} &
A type flag - \obj{delta} (a learned write, invisible at query time until
admitted, kept as provenance once live) or \obj{concrete} (bootstrapped, live
from creation) - the upgrade-only state ladder of \S\ref{sec:graph}, a dedup
key, parent links tying a revision to what it revises, and provenance:
\obj{automated\_onboarding} for bootstrapped objects, the authoring question for
learned ones \\
\bottomrule
\end{tabular}
\end{table}

\begin{figure}[t]
\centering
\IfFileExists{figures/graph.pdf}
  {\includegraphics[width=\textwidth]{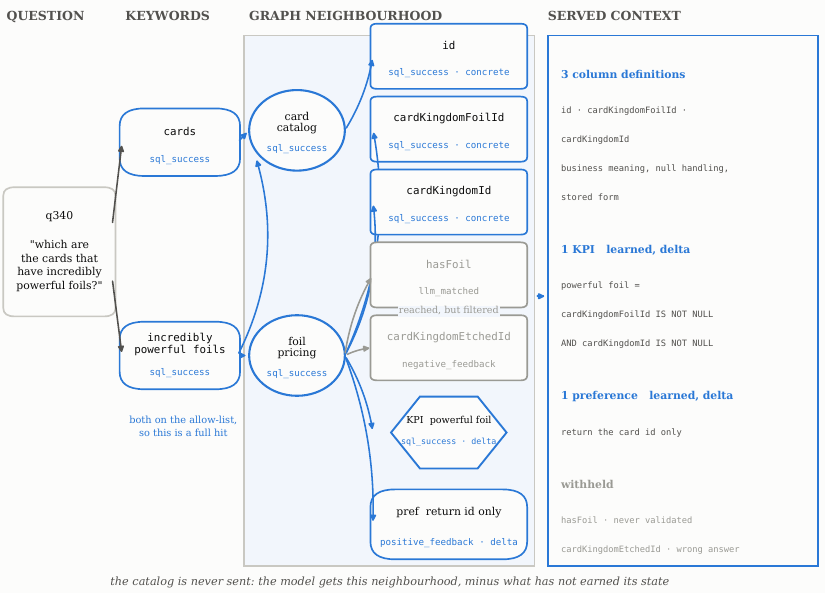}}
  {\fbox{\parbox[c][26mm][c]{0.95\textwidth}{\centering
   question $\rightarrow$ keywords $\rightarrow$ graph neighbourhood $\rightarrow$
   served context}}}
\caption{\textbf{What the graph holds, what it serves, and why the two
differ.} A worked retrieval on question~340; all columns shown belong to
\obj{cards}. Shape marks node type (rounded keyword, elliptical concept,
square column, hexagonal \texttt{kpi}, tagged \texttt{preference}); every
node carries its \obj{semantic\_state} and learned objects their
\obj{delta}/\obj{concrete} provenance. Colour marks one thing: blue is
served, grey is not. The allow-list filters \obj{hasFoil}, reachable but
\obj{llm\_matched} and never validated, and \obj{cardKingdomEtchedId}, which
carried a wrong answer (\obj{negative\_feedback}), while the learned
preference serves under the \texttt{positive\_feedback} written after its
query was verified. That preference is an instance of the projection class
the judge polices (\S\ref{sec:validation}; Table~\ref{tab:judge}), serving
because it passed the gate for its concept's scope. The right block is the
whole of what the SQL agent receives, mixing bootstrapped structure with
learned business meaning.}
\label{fig:graph}
\end{figure}

\paragraph{Node types, semantic state, and the allow-list.}
The graph comprises six node types, structured hierarchically from keywords to
concepts to objects and connected via typed edges (Figure~\ref{fig:graph}).
Crucially, it is constructed dynamically during serving rather than through
onboarding: a node is instantiated only when a query requires it, extracted
keywords being matched against the semantic store and inserted with their
edges. New nodes receive the \texttt{llm\_matched} state which is the lowest
operational state, reflecting that a model-asserted match lacks execution
evidence. Retrieval is a shallow, concept-mediated traversal (keyword
$\rightarrow$ concept $\rightarrow$ object). Outcomes are recorded on the
traversed nodes, so a single execution failure invalidates that routing
pathway for all future queries that depend on it. Each node carries a semantic
state under a strict total ordering:
\begin{center}
\texttt{llm\_matched} $<$ \texttt{initial\_load} $<$ \texttt{positive\_feedback}
$<$ \texttt{sql\_success} $<$ \texttt{sql\_failed} $<$ \texttt{negative\_feedback}
\end{center}
\texttt{sql\_success} and \texttt{sql\_failed} capture execution outcomes;
answer verdicts enter as feedback, with \texttt{negative\_feedback} denoting
SQL that executes cleanly but yields incorrect results (C4,
\S\ref{sec:validation}). The ordering resolves state collisions rather than
reflecting evidential weight. Transitions are strictly monotonic
(upgrade-only) and propagate to parent nodes; since both failure states sit
above both allow-list states, a failed node cannot be returned to
$\mathcal{A}$ by any ordinary outcome, and Eq.~\eqref{eq:recover} is the only
rule that writes a state without taking a maximum. With $\sigma_t(v)$ the
state of node $v$ after outcome $t$, $V(o)$ the nodes engaged by outcome $o$
and $\mathrm{sig}(o)$ its signal, the update is
\begin{equation}
\sigma_{t+1}(v) \;=\; \max\nolimits_{\preceq}\bigl(\sigma_t(v), \mathrm{sig}(o)\bigr)
\qquad \text{for } v \in V(o),
\label{eq:state}
\end{equation}
so incidental successes cannot overwrite recorded failures. Concepts inherit
the maximal failure state of their constituents; a single erroneous column
invalidates its parent concept. Routing uses these states as an
\emph{allow-list}, $\mathcal{A} = \{\texttt{sql\_success},
\texttt{positive\_feedback}\}$: to be served by the graph, a node must have
demonstrated reliability via successful execution or verified positive
feedback. These are criteria on \emph{state}: a node that has never failed
enters $\mathcal{A}$ on its first successful execution, with no reliability
threshold involved. The threshold $\theta = 0.85$ governs only re-entry, which is the
sole exception to monotonicity and is tested on \emph{accumulated}
reliability rather than a single outcome: an invalidated node returns to the
allow-list once the running score $\bar{\rho}(v)$ of Eq.~\eqref{eq:accum},
which weights each successive outcome above the ones before it, reaches
$\theta$ (Eq.~\ref{eq:recover}). A route is thus cheap to earn, immediately
withdrawn on failure, and expensive to earn back.

\textbf{Serving paths and the freshness test.} Write $K(q)$ for the keywords of question $q$, $V(G)$ for the node set of the LCG graph $G$, and $\sigma(k)$ for the semantic state of node $k$. The resolved set is
\begin{equation}
R(q)=\{\,k\in K(q)\;:\;k\in V(G)\ \wedge\ \sigma(k)\in\mathcal{A}\,\},
\label{eq:resolved}
\end{equation}
so $R(q)=K(q)$ is a full hit and $\emptyset\subsetneq R(q)\subsetneq K(q)$ a
partial hit. Because C3 keeps writing new KPIs and preferences underneath the
graph, a freshness test guards the shortcut: the graph is authoritative
exactly when every keyword resolves and nothing in the resolved neighbourhood
postdates the \emph{oldest} keyword's last touch
(Eqs.~\ref{eq:freshness}-\ref{eq:authoritative} below);
the oldest-keyword cutoff prefers redundant work to a stale shortcut. When
the test fails, only the deficit is re-derived and merged with what the graph
held, as the serving-path paragraphs below detail.

\paragraph{Formal definition of freshness.} Let $\tau(k)$ be the time node $k$ was last
updated and $\mathrm{ts}(s)$ an object's \obj{updated\_at} (else
\obj{created\_at}). The query's cutoff is the oldest of its keywords,
\begin{equation}
\tau_{\min}(q)=\min_{k\in K(q)}\tau(k),
\qquad
N(q)=\{\,s\in S(q)\;:\;\mathrm{ts}(s)>\tau_{\min}(q)\,\},
\label{eq:freshness}
\end{equation}
where $S(q)$ are the semantics in the neighbourhood $q$ resolves to, and the
graph is authoritative exactly when
\begin{equation}
R(q)=K(q)\quad\text{and}\quad N(q)=\emptyset .
\label{eq:authoritative}
\end{equation}

\paragraph{Serving behaviour in each case.} When \eqref{eq:authoritative} holds, the
context is read off the graph and nothing is added; the only write is the state
update of \S\ref{sec:graph} (Eq.~\ref{eq:state}), recording how the query fared on the nodes it
used, and during held-out evaluation even that is suppressed: evaluation runs
against frozen state and writes nothing, graph or semantic (\S\ref{sec:e3}). When it fails, either because $K(q)\setminus R(q)\neq\emptyset$ or because
$N(q)\neq\emptyset$, an LLM matching pass runs on the deficit alone, and its
result is \emph{merged} with what the graph already held rather than replacing it:
the older nodes stay, the newly matched semantics are added as nodes with their
edges, so the next question phrased that way finds the union. Edge maintenance
runs in the same step: if C3 has revised a preference the edge is repointed at
the new head, and a superseded head's stale edge is dropped, so a superseded
rule cannot be reached through a route that outlived it (this maintenance
belongs to the serving-path merge, not to C4, which moves state only). The
graph is thus never rebuilt from scratch; it is continuously reconciled
against the semantic store, the two staying in step without either being
regenerated.

\paragraph{Formal specification of the learning writes.} Writes are keyed by
\emph{family} rather than by object, which is what makes a revision a
revision and not a duplicate. Every object $s$ carries
\begin{equation}
\begin{aligned}
\varphi(s)&=\bigl(\mathrm{table}(s),\ \mathrm{type}(s),\ \mathrm{base}(s)\bigr),\\
\mathrm{base}(s)&=\begin{cases}\mathrm{name}(\mathrm{parent}(s)) & \text{if $s$ revises a bootstrap object,}\\ \mathrm{name}(s) & \text{otherwise,}\end{cases}
\end{aligned}
\label{eq:family}
\end{equation}
with $\mathrm{type}\in\{\obj{column},\obj{kpi},\obj{preference}\}$; this is
the \obj{semantic\_family\_key} of Table~\ref{tab:sem-objects}. A proposed
correction $c$ is checked against the learnings already in the store,
\begin{equation}
\mathrm{op}(c)=
\begin{cases}
\text{\emph{update}: append $c$ to $F_{\varphi(c)}$, its head once admitted} & \text{if } F_{\varphi(c)}\neq\emptyset,\\[1pt]
\text{\emph{create}: open $F_{\varphi(c)}=\{c\}$} & \text{otherwise,}
\end{cases}
\label{eq:writeop}
\end{equation}
so a covered fact is revised in place and only a genuinely new fact adds a
rule: two statements under one key never serve at once. A convention
rewritten under a fresh name opens a new family, a leak the judge below only
partly closes (on \obj{card\_games} the 178 learned objects stand on 39
transfer families), and the residual regressions of \S\ref{sec:e3} trace to
single heads. Writing $F_k=\{s\in\mathcal{S}:\varphi(s)=k\}$ for the family
under key $k$, exactly one member is live at a time, the most recently
written \emph{admitted} member:
\begin{equation}
\mathrm{head}(F_k)=\operatorname*{arg\,max}_{s\in F_k,\ \mathrm{admitted}(s)}\ \mathrm{ts}(s),
\qquad
\mathcal{S}^{\ast}=\{\,\mathrm{head}(F_k)\ :\ k\,\},
\label{eq:head}
\end{equation}
a family with no admitted member serving nothing, so a rejected proposal
never becomes head and cannot displace the rule it tried to revise.
Retrieval reads $\mathcal{S}^{\ast}$ and never $\mathcal{S}$: a rule becomes
live by becoming its family's head and inert by being superseded, not
deleted. The $\mathrm{admitted}(\cdot)$ predicate is the approval gate of
\S\ref{sec:validation}, a proposed rule reviewed before it serves, by a human
in deployment and an LLM judge mimicking that reviewer here; it exists for
what family keying cannot catch: a question asking for a \emph{name} whose
expected answer carries only the id would, learned faithfully, teach a
projection rule that breaks other, correct queries. Given a verified outcome
$(q,\ \mathrm{SQL}_q,\ C_q)$, where $C_q$ is the context that was served, the
agent emits writes $W(q)$ and the store grows by union:
\begin{equation}
\mathcal{S}_{t+1}=\mathcal{S}_t\ \cup\ W(q),
\qquad
\mathrm{parent}(w)=\mathrm{head}\bigl(F_{\varphi(w)}\bigr)\ \text{where one exists.}
\label{eq:update}
\end{equation}
The store is therefore \emph{append-only}, $\mathcal{S}_t\subseteq
\mathcal{S}_{t+1}$ for all $t$, an update changing only which element of a
family is its head: rollback is exact, since restoring a prior head restores
the prior served knowledge base, and forgetting in \S\ref{sec:e2} is
\emph{negative transfer} rather than overwriting; nothing is lost, a worse
rule is merely in front.

\paragraph{Outcome reliability scoring.} Graph invalidation and its reversal are driven
by a reliability score $\rho\in[0,1]$, a weighted mean over five signals so none
decides alone. With $\mathcal{I}$ the signals applicable to a query,
\begin{equation}
\rho = \frac{\sum_{i\in\mathcal{I}} w_i s_i}{\sum_{i\in\mathcal{I}} w_i},
\qquad
w_{\mathrm{val}}{=}0.45,\ w_{\mathrm{exec}}{=}0.20,\ w_{\mathrm{ex}}{=}0.15,\
w_{\mathrm{gen}}{=}0.12,\ w_{\mathrm{db}}{=}0.08
\label{eq:rel}
\end{equation}
for validation, execution, example support, generation efficiency and database
routing. Normalising by $\sum_{i\in\mathcal{I}} w_i$ rather than $1$ makes
inapplicable signals \emph{masked} rather than scored zero. Two of the five are
tied directly to the states of \S\ref{sec:graph} and neither is a judgement. The
execution term is a hard fact: $0$ if the SQL errored (\obj{sql\_failed}), $0.5$
if it ran and returned no rows, $1$ if it returned rows (\obj{sql\_success}); the
middle case matters, because the commonest way to return nothing is to filter on
a value that does not exist in the form the query assumed. The validation term is
binary, $1$ on a verified-correct answer (\obj{positive\_feedback}) and $0$ on a
verified-wrong one (\obj{negative\_feedback}), masked where no verdict exists;
carrying the largest weight, it is what makes a cleanly-executing query that
returns the wrong answer score below one that errors outright. The remaining
three are LLM-judged over the answer path, generation efficiency scoring the
attempts and failures a question consumed before it resolved. Validation was
available throughout these runs; where it is not, $\rho\leftarrow\min(\rho,0.65)$,
which matters below.

\paragraph{Accumulation over a node's outcomes.} A single $\rho(o)$ scores an
outcome, not a route. Each node instead carries a running score over the outcomes
that engaged it, exponentially weighted with $\alpha=0.6$ so that recent evidence
dominates without older evidence being discarded:
\begin{equation}
\bar{\rho}_{t+1}(v) \;=\; (1-\alpha)\,\bar{\rho}_{t}(v) \;+\; \alpha\,\rho(o),
\qquad v \in V(o), \quad \bar{\rho}_{0}(v)=0,
\label{eq:accum}
\end{equation}
so the zero start is a conservative prior rather than a score: a node's first
outcome carries it only to $\alpha\,\rho(o)$, and after $n$ outcomes the
$i$-th carries weight $\alpha(1-\alpha)^{n-i}$, the residual $(1-\alpha)^{n}$
resting on the prior, after three, $(0.096, 0.24, 0.6)$ on the outcomes and
$0.064$ on the prior, so no single outcome, however good, clears
$\theta$ alone. Two
properties follow and both are load-bearing. A node is judged on repetition
rather than on one query: an incidental failure by a node with a long verified
history costs it less than the same failure costs a node that has never
succeeded, which is as much credit assignment as a shared-node design admits.
And the number of verified successes needed to undo a failure is analytic rather
than stipulated, $k \ge \ln\bigl((1-\theta)/(1-\bar{\rho})\bigr)/\ln(1-\alpha)$:
three consecutive outcomes at $\rho(o)=1$ carry a node from $0$ to $0.936$, while
$\alpha$ is the knob that sets how expensive re-earning a route is.

\paragraph{Asymmetry between invalidation and recovery.} A failed execution
marks every keyword and concept node that supplied the failing context
\obj{sql\_failed}; an answer verified wrong marks them
\obj{negative\_feedback}; and because nodes are shared (\S\ref{sec:graph}),
either mark withdraws the route from every question that would have used it,
not only the one that failed. Both states lie outside the allow-list and
above $\obj{sql\_success}$ in the upgrade-only order, so the marks are
\emph{sticky}. Reversal bypasses the ordering but demands more, of the
accumulated score rather than one outcome:
\begin{equation}
\begin{aligned}
&\sigma_t(v)\in\{\obj{sql\_failed},\obj{negative\_feedback}\}\ \wedge\ \mathrm{sig}(o)\in\mathcal{A}\ \wedge\ \bar{\rho}_{t+1}(v)\ge\theta\\
&\qquad\Longrightarrow\ \sigma_{t+1}(v)=\mathrm{sig}(o),
\end{aligned}
\label{eq:recover}
\end{equation}
with $\theta{=}0.85$ and $\bar{\rho}$ from Eq.~\eqref{eq:accum}. The
thresholds interact: $\rho(o)$ is capped at $0.65$ without validation, and a
weighted mean of capped outcomes stays below $0.65<\theta$, so recovery is
\emph{unreachable} for a node whose successes were never independently
validated. A revived node returns as \obj{sql\_success} when the reviving
signal is the execution and as \obj{positive\_feedback} when it is the
post-learning feedback; that validation is itself LLM-judged, so the recovery
gate rests on the same self-assessed plausibility
signal~\citep{kadavath2022know} that \S\ref{sec:e2} argues admission should
move beyond; grounding both in measured execution evidence is the same open
problem. The asymmetry is deliberate: invalidation withdraws a \emph{routing
shortcut}, not knowledge; the definitions remain in the semantic layer and an
unresolved question falls back to ordinary context assembly, so wrongly
invalidating costs a slower answer, wrongly retaining an incorrect one.

\paragraph{What engages an invalidated node.} Invalidation withdraws a node
from the \emph{served} context, not from the graph: its edges and accumulated
score persist (Figure~\ref{fig:graph}, \emph{reached, but filtered}). $V(o)$
is the nodes an outcome's serving stood on, and a filtered node re-enters it
only when the fallback matching pass independently re-selects the same
underlying object: the merge re-lays the route, the object is served the
expensive way, and the outcome engages the node whose content it used. The
successes driving $\bar{\rho}$ toward $\theta$ in Eq.~\eqref{eq:recover} are
therefore successes of queries that actually consumed the node's semantics,
never of queries that merely touched its keyword: a withdrawn route the
matching pass stops selecting stays down; one whose semantics keep being
re-derived and keep verifying earns its shortcut back.

\paragraph{Cold start from query history.} This paragraph is deployment
design, not anything the experiments run; Exp~3's stream is the live phase
replicated under measurement. The learning step is indifferent to where an
outcome comes from, so one mechanism serves both phases of a deployment: given
a corpus of past queries whose SQL ran, was accepted, and whose question is
recoverable, it replays them as though just asked, the known-good SQL
supplying the verified outcome directly with no retry loop, warm-starting the
knowledge base without anyone authoring a definition; given live traffic it
does the same on each new question. These are not two modes: history is
processed by exactly the serving path of Figure~\ref{fig:pipeline}, and the
knowledge base it leaves behind is the one the next live question inherits.
The organization contributes its data and its past use of it; it is never
asked for a semantic layer.

\paragraph{Scoring protocol.} BIRD execution accuracy: gold SQL on SQLite, the
prediction on a MySQL copy with identical data, row sets compared as sets. We
use a 300\,s timeout rather than BIRD's default 30\,s because several correct queries need longer (one returns exactly the gold's 546 rows in ${\sim}40$\,s), so any score at the default is a lower bound rather than a
measurement. One shared function scores every run reported here.

\begin{table}[!t]
\centering
\caption{\textbf{The retrieval baseline across $k$.} Held-out correct with the $k$ most similar training questions and their gold SQL in the prompt, over the raw schema (row (r) of Table~\ref{tab:learned} at $k{=}5$) and over the bootstrapped catalog (row (s)). $k{=}5$, run first, is within one question of the sweep's best in both arms, and accuracy falls at $k{=}10$: more retrieved examples do not close the gap to the learned base's 53.}
\label{tab:ksweep}
\small
\begin{tabular}{@{}lccc ccc@{}}
\toprule
& \multicolumn{3}{c}{\textbf{raw schema $+\,k$}} & \multicolumn{3}{c}{\textbf{bootstrapped $+\,k$}} \\
\cmidrule(lr){2-4}\cmidrule(l){5-7}
$k$ & 3 & 5 & 10 & 3 & 5 & 10 \\
\midrule
\obj{card\_games} (57) & 10 & 13 & 12 & 12 & 12 & 10 \\
\obj{toxicology} (44) & \ 9 & \ 8 & \ 8 & \ 9 & \ 7 & \ 6 \\
\obj{thrombosis\_prediction} (45) & 12 & 14 & 12 & 13 & 14 & 11 \\
\midrule
pooled (146) & 31 & \textbf{35} & 32 & 34 & \textbf{33} & 27 \\
\bottomrule
\end{tabular}
\end{table}

\paragraph{The few-shot retrieval baseline.} Rows (r) and (s) of
Table~\ref{tab:base}: two prompt arms at $k{=}5$ (the sweep over $k\in\{3,5,10\}$ in Table~\ref{tab:ksweep} varies only $k$), the raw schema $+$ the five examples
(r) and the bootstrapped catalog $+$ the same five (s), consuming identical
rankings. The corpus holds one document per training question
(the question text); the query is the test question; nothing from the test
split is indexed, asserted per retrieved id. \emph{BM25}: Lucene via
\obj{pyserini}~1.6.0~\citep{lin2021pyserini} rather than \obj{rank\_bm25}, because at 100--350
documents the analyzer dominates the scoring function; the stock
\obj{EnglishAnalyzer} (lowercase, punctuation stripped, English stoplist,
Porter stemming, the stemmer Lucene ships, not Krovetz) with no min-df
pruning, since \obj{Coldsnap}, \obj{TR000} and \obj{Portuguese (Brazil)} are
exactly the discriminative terms. Both parameterisations run: the Lucene
default $k_1{=}1.2$, $b{=}0.75$ and the MS-MARCO-tuned $k_1{=}0.9$,
$b{=}0.4$. \emph{Dense}: \obj{intfloat/e5-large-v2}~\citep{wang2022e5} with its required
asymmetric prefixes (\obj{query:}\,/\,\obj{passage:}), cosine over
L2-normalised vectors, exact search, no ANN at this corpus size.
\emph{Hybrid}: reciprocal rank fusion~\citep{cormack2009rrf} with $k_{\mathrm{RRF}}{=}60$,
$\mathrm{score}(d)=\sum 1/(60+\mathrm{rank}(d))$, rank-based, so no score
normalisation. \emph{Determinism}: seed 20260820, every ranking stably
sorted on $(-\mathrm{score},\,\mathrm{question\_id})$; rankings are built
once into a static JSON under an isolated environment (\obj{pyserini} needs
a JVM, the dense model needs \obj{torch}) and read by the runner, so the
artifact is inspectable and the experiment environment stays pinned.
\emph{Oracle advantage, declared}: the BM25 variant fused into the hybrid is
chosen using test labels, recorded in the artifact as
\obj{bm25\_selected\_by}; the semantic-layer arms got no equivalent choice.

\FloatBarrier
\section{Additional results and protocol}
\label{app:results}

\begin{table}[H]
\centering
\caption{\textbf{The admission judge is load-bearing.} Condition (e) re-run with the judge disabled, every proposed learning admitted; both arms learn from the training questions only. KPI and pref columns count the learned rules each store holds: the judge admits fewer rules (pooled, 410 against 488 KPIs, 354 against 421 prefs) and scores higher everywhere, pooled held-out falling 53 to 29 of 146 without it and train 269 to 207 of 353; filtering, not volume, carries the gain. It exists for the noisy-supervision class of \S\ref{sec:validation} (the name-asked, id-answered projection rule); in enterprise deployment the reviewer is the human approval already on the write path.}
\label{tab:judge}
\small
\setlength{\tabcolsep}{4.5pt}
\begin{tabular}{@{}lcccc cccc@{}}
\toprule
& \multicolumn{4}{c}{\textbf{with judge}} & \multicolumn{4}{c}{\textbf{without judge}} \\
\cmidrule(lr){2-5}\cmidrule(l){6-9}
& train & h.o. & KPIs & prefs & train & h.o. & KPIs & prefs \\
\midrule
\obj{card\_games} & 105 & 22 & \ 95 & \ 83 & 74 & 10 & 113 & 100 \\
\obj{toxicology} & \ 87 & 12 & 138 & \ 88 & 71 & \ 7 & 166 & 112 \\
\obj{thrombosis\_prediction} & \ 77 & 19 & 177 & 183 & 62 & 12 & 209 & 209 \\
\midrule
pooled & 269 & \textbf{53} & 410 & 354 & 207 & \textbf{29} & 488 & 421 \\
\bottomrule
\end{tabular}
\end{table}

\begin{table}[H]
\centering
\caption{\textbf{The retry loop, priced.} Questions wrong on the first pass of Exp~2's one-shot learning enter the answer-matched retry loop of \S\ref{sec:learning}, capped at ten attempts; a question not fixed within the cap is not learned from. Every fixed question needed at most eight attempts.}
\label{tab:retry}
\small
\begin{tabular}{@{}lccc@{}}
\toprule
& \textbf{\obj{card\_games}} & \textbf{\obj{toxicology}} & \textbf{\obj{thrombosis\_pred.}} \\
\midrule
eligible (wrong first pass) & 63 & 22 & 93 \\
fix rate & 85.7\% (54) & 86.4\% (19) & 80.6\% (75) \\
mean attempts to fix & 3.82 & 2.90 & 4.11 \\
max attempts to fix & \ 6 & \ 5 & \ 8 \\
\bottomrule
\end{tabular}
\end{table}

\textbf{The remaining gap is the own-example effect.} The oracle on \obj{card\_games}, learning from all 191 including held-out, reaches 38/57: of the 35 questions the transfer condition misses, roughly half are
recoverable in principle and the rest are not learnable by this mechanism at all; the honest target for future work. The oracle's train score is \emph{lower} than (e)'s on every database, 101 against 105, 86 against 87, 74 against 77: the held-out questions' rules made previously-correct train questions wrong, \obj{toxicology}'s one-question gap we do not interpret; the four- and three-question gaps on the other two databases are the evidence.

\paragraph{Checkpoint protocol.} At each checkpoint, at $t{=}20$ for example,
the batch's 20 questions are run live and the graph is built as they are
answered: nodes are created \obj{llm\_matched}, promoted to
\obj{sql\_success} by execution, or marked \obj{sql\_failed}. The answers are
then evaluated: a question right on the first pass is learned from as it
stands; a wrong one is marked \obj{negative\_feedback}, withdrawing its nodes
even where the SQL ran cleanly, and enters the retry loop, each still-wrong
attempt renewing the mark, until the system's own SQL scores correct, then is
learned from too. Both then receive positive feedback, written after the
learning: no new semantics attach in this pass, only the states of the nodes
each question stood on move, and the next batch's serving links the new rules
into the graph. Nodes marked \obj{sql\_failed} or \obj{negative\_feedback}
stay off the allow-list until a later success clears the recovery threshold
(Eq.~\ref{eq:recover}). Correction is the system re-prompted to explore the
data and try again, scored by answer matching, the gold SQL neither shown nor
used, the system receiving only each attempt's verdict (\S\ref{sec:learning}).

\begin{figure}[t]
\centering
\begin{minipage}[c]{0.46\textwidth}
\centering
\IfFileExists{figures/curve.pdf}
  {\includegraphics[width=\linewidth]{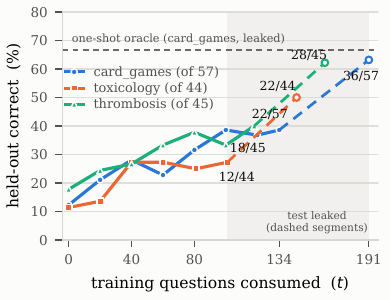}}
  {\fbox{\parbox[c][40mm][c]{0.92\linewidth}{\centering held-out accuracy against $t$}}}
\end{minipage}\hfill
\begin{minipage}[c]{0.50\textwidth}
\caption{\textbf{Parity, on three databases.} Held-out accuracy per checkpoint: \obj{card\_games} ends at 22/57 and \obj{toxicology} at 12/44, each exactly one-shot (row (e), Table~\ref{tab:learned}); \obj{thrombosis\_prediction} at 18/45 against 19/45. Gray: the leaked \obj{card\_games} oracle (38/57). In the shaded region \textbf{the test set is leaked}: the dashed continuations are upper bounds, climbing to 36/57, 22/44 and 28/45.}
\label{fig:curve}
\end{minipage}
\end{figure}

\textbf{Leaked continuations as upper bounds.} Continuing the
\obj{card\_games} stream past $t{=}134$ through the 57 held-out questions
\emph{leaks the test set}: every number past that point is an upper bound,
marked in Table~\ref{tab:e3full} and Figure~\ref{fig:curve}. Read as that bound, the leaked streams reach \textbf{36/57}, 22/44 and 28/45 against one-shot oracles of 38/57, 22/44 and 27/45 (row (g), Table~\ref{tab:base}): the same shape, read as the same upper bound.

\begin{figure}[t]
\centering
\begin{minipage}[c]{0.46\textwidth}
\centering
\IfFileExists{figures/cost.pdf}
  {\includegraphics[width=\linewidth]{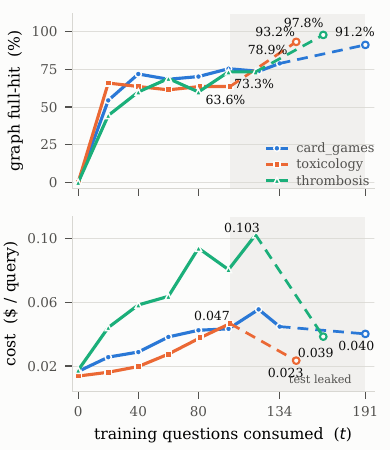}}
  {\fbox{\parbox[c][40mm][c]{0.92\linewidth}{\centering coverage and cost}}}
\end{minipage}\hfill
\begin{minipage}[c]{0.50\textwidth}
\caption{\textbf{Coverage is earned; cost follows what the graph can route.} \emph{Top:} graph full-hit rate. \emph{Bottom:} per-query cost. \obj{card\_games}' coverage is still rising at $t{=}134$ and its cost falls at that final checkpoint; \obj{toxicology} plateaus near 63.6\% with cost rising, and \obj{thrombosis\_prediction} climbs to 73.3\% at \$0.1025 per query. In the shaded leaked continuations cost falls as coverage nears totality, though there the graph holds routes for the evaluated questions themselves. Median latency is in Table~\ref{tab:e3full}.}
\label{fig:cost}
\end{minipage}
\end{figure}

\begin{figure}[t]
\centering
\IfFileExists{figures/hitpaths.pdf}
  {\includegraphics[width=\textwidth]{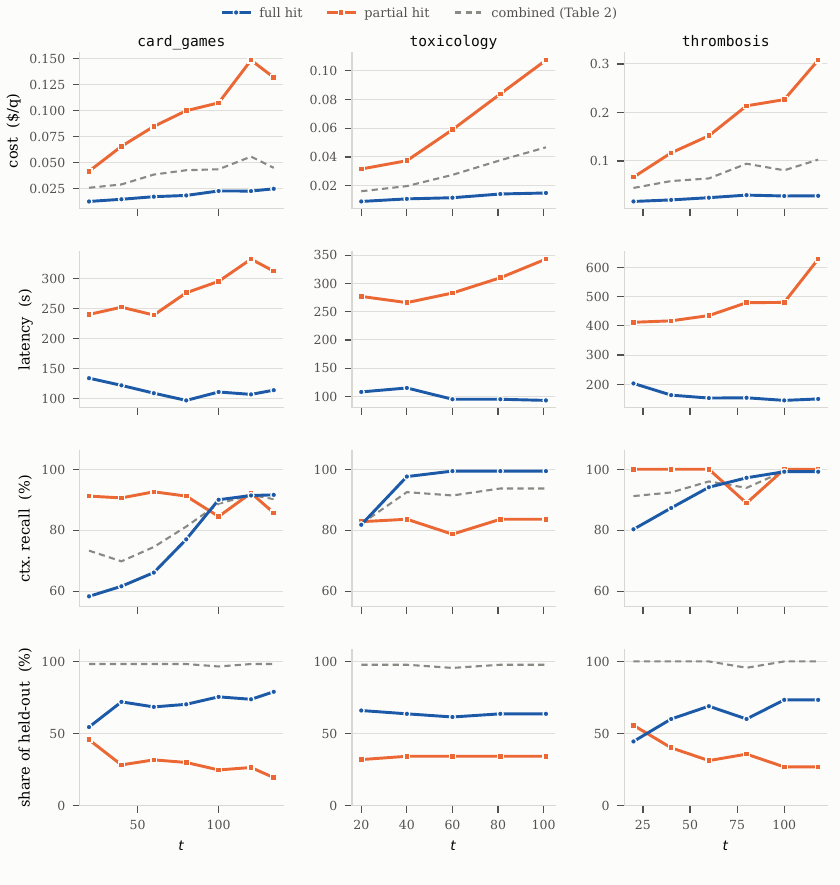}}
  {\fbox{\parbox[c][30mm][c]{0.95\textwidth}{\centering full vs partial: cost, latency, share}}}
\caption{\textbf{Full hits are cheap; partial hits carry the growth.} Per-checkpoint cost, latency, context recall and serving-path share, split full against partial (data in Table~\ref{tab:e3full}). The combined line is Table~\ref{tab:e3full}'s blended figure on the cost, recall and share panels (share: at-least-partial); its latency is a median rather than the mean of the group figures, so the latency panels carry the two paths. Full-hit recall climbs toward and past the partial path as the graph matures: the served neighbourhood becomes as complete as the fallback pass's.}
\label{fig:hitpaths}
\end{figure}

\textbf{The dips coincide with cost, on one run.} The accuracy dips are within
the single-evaluation noise noted in the \limref, so this describes one run,
consistent with the mechanism rather than a test of it. On the
\obj{card\_games} stream (Table~\ref{tab:e3full}), the two steps where
held-out accuracy falls, $t{=}40 \to 60$ and $t{=}100 \to 120$, are the only
two where graph full-hit coverage \emph{also} falls, by 3.5 and 1.7 points.
The first two checkpoints build the graph from nothing and cost tracks that
construction; across the five steps after it exists ($t \ge 40$), the two
accuracy dips carry the two largest cost increases, $+33\%$ and $+28\%$ per
query, the two largest coverage gains, both $+5.2$ points, carry the two
smallest cost changes, $+2\%$ and $-19\%$, and the inverse ordering of cost
change against coverage change is exact across all five.

\textbf{Cost growth and the saturation hypothesis.} Per-query cost rises on
every unleaked stream, ending $2.7$--$5.9\times$ above its start (\$0.0166 to
\$0.0448, \$0.0137 to \$0.0467, \$0.0174 to \$0.1025). The rise comes from
hundreds of learned rules entering the served context, charged only to the
questions the graph cannot yet route: at every measured checkpoint on every
stream, a question served entirely from the graph costs $3.3$--$11\times$
less than one that falls back to partial matching and answers
$1.8$--$4.2\times$ faster (Table~\ref{tab:e3full};
Figure~\ref{fig:hitpaths}). That cost saturates once coverage is high is the
design's prediction (\S\ref{sec:graph}) and at these stream lengths remains a
hypothesis: its support is the only two unleaked steps where cost falls,
\obj{card\_games}' final checkpoint ($-19\%$) and
\obj{thrombosis\_prediction}'s $t{=}100$ ($-14\%$), both coinciding with
full-hit jumps ($+5.2$ and $+13.3$ points), and none on \obj{toxicology},
whose coverage plateaus. The leaked continuations point the same way: as full
hits reach 91--98\%, cost falls on every stream (Table~\ref{tab:e3full};
Figure~\ref{fig:cost}), though there the graph holds routes for the evaluated
questions themselves, so the fall is an upper bound on what coverage buys.

The pattern is the mechanism of \S\ref{sec:graph} priced: a batch that
produces failures marks the nodes that routed the failing context
\obj{sql\_failed}, so coverage falls because the graph serves fewer validated
shortcuts, not fewer nodes, and every affected question falls back to the
full matching pass, where re-derivation is what costs money. The jump is not
the knowledge base growing, which grew by roughly as much in the batches
either side; it is the same questions answered the expensive way. The final
checkpoint is the symmetric case: coverage recovers to 78.9\%, its high-water
mark, as invalidated nodes earn their way back, and cost falls $-19\%$, the
one place in the stream where a larger knowledge base is served more cheaply
than a smaller one. Serving cost thus tracks what the graph can route, not
what the store contains, appearing here as a single point rather than a
trend. The full three-way coupling of accuracy dip, coverage fall and cost
rise appears on \obj{card\_games} only, since on the replications coverage
can rise through an accuracy dip; what survives on all three streams is that
every coverage fall carries a cost rise (\S\ref{sec:e3}).

\begin{table}[tp]
\centering
\caption{\textbf{Exp~3, every checkpoint.} The same held-out questions evaluated at every checkpoint against frozen state; the endpoints are Table~\ref{tab:e3}. Latency in the \emph{all} column is the median over held-out questions; the full/partial columns are means within each serving path, so each $\times$-factor in the text compares like with like. $^{\dagger}$: the final store re-evaluated with the graph disabled; $^{\ast}$: the stream continued through its own held-out questions, so the test set is leaked and the row is an upper bound only. Where measured, \$/query, latency and context recall are split by serving path, full graph hit against partial: a full hit costs $3.3$--$11\times$ less and answers $1.8$--$4.2\times$ faster at every measured checkpoint (Figure~\ref{fig:hitpaths}).}
\label{tab:e3full}
\scriptsize
\setlength{\tabcolsep}{3pt}
\begin{tabular}{@{}rrrrrrrrrr@{}}
\toprule
& \multicolumn{1}{c}{\textbf{Held-out}} & \multicolumn{2}{c}{\textbf{\$/query}} & \multicolumn{2}{c}{\textbf{latency (s)}} & \multicolumn{2}{c}{\textbf{ctx. recall}} & \multicolumn{2}{c}{\textbf{Graph hit}} \\
\cmidrule(lr){2-2}\cmidrule(lr){3-4}\cmidrule(lr){5-6}\cmidrule(lr){7-8}\cmidrule(l){9-10}
$t$ & acc. & all & full\,/\,partial & all & full\,/\,partial & all & full\,/\,partial & full & partial \\
\midrule
\multicolumn{10}{@{}l}{\obj{card\_games}: 57 held-out, 134 training questions} \\
0 & \ 7 \ (12.3\%) & 0.0166 & -- & 165.1 & -- & 87.0\% & -- & \ 0.0\% & \ 0.0\% \\
20 & 12 \ (21.1\%) & 0.0256 & 0.0124\,/\,0.0415 & 177.1 & 134\,/\,240 & 73.3\% & 58.3\,/\,91.2\% & 54.4\% & 98.2\% \\
40 & 16 \ (28.1\%) & 0.0288 & 0.0145\,/\,0.0655 & 137.2 & 122\,/\,252 & 69.8\% & 61.6\,/\,90.6\% & 71.9\% & 98.2\% \\
60 & 13 \ (22.8\%) & 0.0383 & 0.0170\,/\,0.0847 & 128.3 & 109\,/\,239 & 74.5\% & 66.1\,/\,92.6\% & 68.4\% & 98.2\% \\
80 & 18 \ (31.6\%) & 0.0425 & 0.0182\,/\,0.0998 & 109.6 & \ 97\,/\,276 & 81.2\% & 77.0\,/\,91.2\% & 70.2\% & 98.2\% \\
100 & \textbf{22 \ (38.6\%)} & 0.0434 & 0.0225\,/\,0.1074 & 122.3 & 111\,/\,295 & 88.7\% & 90.0\,/\,84.5\% & 75.4\% & 96.5\% \\
120 & 21 \ (36.8\%) & 0.0556 & 0.0224\,/\,0.1485 & \ 98.0 & 107\,/\,332 & 91.6\% & 91.4\,/\,92.2\% & 73.7\% & 98.2\% \\
134 & \textbf{22 \ (38.6\%)} & 0.0448 & 0.0246\,/\,0.1319 & 110.0 & 114\,/\,312 & 90.2\% & 91.6\,/\,85.7\% & \textbf{78.9\%} & 98.2\% \\
\addlinespace[1pt]
134$^{\dagger}$ & 23 \ (40.4\%) & 0.1477 & -- & 373 & -- & 95.6\% & -- & -- & -- \\
\addlinespace[1pt]
191$^{\ast}$ & \textbf{36 \ (63.2\%)} & 0.0402 & 0.0278\,/\,0.1695 & 182.8 & 195\,/\,407 & 97.0\% & 98.7\,/\,80.0\% & \textbf{91.2\%} & 98.2\% \\
\midrule
\multicolumn{10}{@{}l}{\obj{toxicology}: 44 held-out, 101 training questions} \\
0 & \ 5 \ (11.4\%) & 0.0137 & -- & 172 & -- & 88.1\% & -- & \ 0.0\% & \ 0.0\% \\
20 & \ 6 \ (13.6\%) & 0.0161 & 0.0090\,/\,0.0316 & 161 & 108\,/\,277 & 82.1\% & 81.8\,/\,82.8\% & 65.9\% & 97.7\% \\
40 & 12 \ (27.3\%) & 0.0197 & 0.0108\,/\,0.0374 & 165 & 115\,/\,266 & 92.5\% & 97.6\,/\,83.6\% & 63.6\% & 97.7\% \\
60 & 12 \ (27.3\%) & 0.0275 & 0.0116\,/\,0.0590 & 158 & \ 95\,/\,283 & 91.4\% & 99.4\,/\,78.7\% & 61.4\% & 95.5\% \\
81 & 11 \ (25.0\%) & 0.0377 & 0.0142\,/\,0.0839 & 167 & \ 95\,/\,310 & 93.7\% & 99.4\,/\,83.6\% & 63.6\% & 97.7\% \\
101 & \textbf{12 \ (27.3\%)} & 0.0467 & 0.0149\,/\,0.1074 & 181 & \ 93\,/\,343 & 93.7\% & 99.4\,/\,83.6\% & 63.6\% & 97.7\% \\
\addlinespace[1pt]
101$^{\dagger}$ & 11 \ (25.0\%) & 0.1084 & -- & 326 & -- & 98.9\% & -- & -- & -- \\
\addlinespace[1pt]
145$^{\ast}$ & \textbf{22 \ (50.0\%)} & 0.0234 & 0.0180\,/\,0.0965 & \ 98 & 101\,/\,250 & 94.1\% & 96.5\,/\,60.0\% & \textbf{93.2\%} & 97.7\% \\
\midrule
\multicolumn{10}{@{}l}{\obj{thrombosis\_prediction}: 45 held-out, 118 training questions} \\
0 & \ 8 \ (17.8\%) & 0.0174 & -- & 212 & -- & 92.7\% & -- & \ 0.0\% & \ 0.0\% \\
20 & 11 \ (24.4\%) & 0.0442 & 0.0163\,/\,0.0664 & 319 & 203\,/\,412 & 91.2\% & 80.3\,/\,100.0\% & 44.4\% & 100.0\% \\
40 & 12 \ (26.7\%) & 0.0584 & 0.0195\,/\,0.1166 & 265 & 163\,/\,417 & 92.4\% & 87.3\,/\,100.0\% & 60.0\% & 100.0\% \\
60 & 15 \ (33.3\%) & 0.0638 & 0.0241\,/\,0.1518 & 240 & 153\,/\,435 & 96.0\% & 94.1\,/\,100.0\% & 68.9\% & 100.0\% \\
80 & 17 \ (37.8\%) & 0.0941 & 0.0294\,/\,0.2136 & 269 & 154\,/\,479 & 93.9\% & 97.2\,/\,88.9\% & 60.0\% & 95.6\% \\
100 & 15 \ (33.3\%) & 0.0806 & 0.0276\,/\,0.2264 & 234 & 145\,/\,480 & 99.4\% & 99.2\,/\,100.0\% & 73.3\% & 100.0\% \\
118 & \textbf{18 \ (40.0\%)} & 0.1025 & 0.0279\,/\,0.3078 & 277 & 150\,/\,628 & 99.4\% & 99.2\,/\,100.0\% & 73.3\% & 100.0\% \\
\addlinespace[1pt]
118$^{\dagger}$ & 19 \ (42.2\%) & 0.2755 & -- & 539 & -- & 95.6\% & -- & -- & -- \\
\addlinespace[1pt]
163$^{\ast}$ & \textbf{28 \ (62.2\%)} & 0.0385 & 0.0352\,/\,0.1842 & 126 & 120\,/\,391 & 99.4\% & 99.4\,/\,100.0\% & \textbf{97.8\%} & 100.0\% \\
\bottomrule
\end{tabular}
\end{table}

\begin{table}[t]
\centering
\caption{\textbf{Invalidated questions mostly recover.} A break event is a held-out question right at one checkpoint and wrong at the next; it recovers if right again at any later checkpoint; final-checkpoint breaks are censored (\S\ref{sec:e3}). Stricter readings: recovery within a single checkpoint is 16/41 (39\%) pooled, and roughly 20 of 33 distinct questions ever recover. All counts come from single evaluations, so a verdict flip registers as a break and as a recovery alike; the rates are upper bounds on what invalidation accounts for (\limref).}
\label{tab:recovery}
\small
\begin{tabular}{@{}lccc@{}}
\toprule
& \textbf{break events} & \textbf{recovered} & \textbf{rate} \\
\midrule
\obj{card\_games} & 19 & 10 & 53\% \\
\obj{toxicology} & \ 7 & \ 5 & 71\% \\
\obj{thrombosis\_prediction} & 15 & 10 & 67\% \\
\midrule
pooled & 41 & 25 & \textbf{61\%} \\
\bottomrule
\end{tabular}
\end{table}

\section{Extended discussion}
\label{app:discussion}

\textbf{Contents of the supplied-knowledge channel.} Supplied knowledge enters as
its own labelled block and governs where it conflicts with a learned rule. What
varies is what enters the channel: in deployment, whatever the organization already trusts; in these experiments, only what BIRD
supplies, and only in the four conditions marked $\ddagger$.

\begin{table}[t]
\centering
\caption{\textbf{Learning the teachers does not regress the independents.} The 64, 42 and 16 training questions whose facts no held-out question needs (\S\ref{sec:split}), scored against the bootstrapped catalog ($t{=}0$) and against the final learned store of row (e), Table~\ref{tab:learned}, both with the graph off. Independents are learned from like every training question, so the column bounds interference from teacher-targeted rules net of each question's own-example fit; it does not isolate it.}
\label{tab:independents}
\small
\begin{tabular}{@{}lcccc@{}}
\toprule
& teachers $n$ & independents $n$ & $t{=}0$ (bootstrapped) & learned (final) \\
\midrule
\obj{card\_games} & 70 & 64 & 29 & 46 \\
\obj{toxicology} & 59 & 42 & 29 & 33 \\
\obj{thrombosis\_prediction} & 102 & 16 & \ 5 & \ 9 \\
\bottomrule
\end{tabular}
\end{table}

\textbf{The same result as backward transfer.} In continual-learning terms this is backward transfer~\citep{lopezpaz2017gem}: with the training set as the first task and the held-out set as the second, BWT is the change in train accuracy from row (e) to row (g) of Table~\ref{tab:base}, $-4/134$, $-1/101$ and $-3/118$ ($-3.0$, $-1.0$ and $-2.5$ points; pooled $-8/353$, $-2.3$ points), at matched supervision. Two caveats. Per-checkpoint BWT over the Exp~3 streams is not available, because train accuracy was not re-measured at checkpoints. And with BIRD's hints supplied the sign flips (row (h) train 281 against row (f) 273, $+8$), so the negative transfer reported here is a property of the hint-free, deployed setting, not of the store in every configuration.

\textbf{Curated domain knowledge is not sufficient, and not uniformly necessary.} Rows (c)--(f) of Table~\ref{tab:base} price BIRD's own premise. On held-out, where nothing was learned from, curated hints over a raw schema reach 38 of 146 (26.0\%); the knowledge base the system wrote from its own outcomes, never seeing an evidence line, reaches 53 (36.3\%), $+10.3$ points; together they reach 60 (row (f)). The hints' contribution is uneven: on \obj{toxicology} they do not help ((c) 82 against (b) 83 of 145) and with the catalog hurt ((d) 79), and pooled the bootstrapped catalog adds nothing over them ((d) 37 against (c) 38); only (d)'s gap on \obj{toxicology} is large enough to interpret. A hint resolves one linking decision; the residue (which columns bear on the query, what a null means, how a value is stored) is \emph{schema linking}~\citep{pourreza2023dinsql,talaei2024chess,wang2025linkalign}, the sharpest distinction from retrieval-augmented prompting~\citep{lewis2020rag}: the right text is necessary, but knowing how to use it is worth more. Rows (r) and (s) price this directly, retrieving the very training questions reaching at best 35 of 146 against the learned base's 53 (Table~\ref{tab:ksweep}). Over all 499 questions the base reaches 64.5\%, 17.6 points above row (c), but that column includes the 353 training questions the base learned from, where the gap ($+20.7$ points) measures fit to its own supervision as much as transfer; the split of \S\ref{sec:split} exists to remove that confound, so the transfer claim rests on held-out. Once a question has been learned from, curated knowledge adds little: (g) and (h) match within noise on \obj{card\_games} (38 against 37), while hints add three held-out questions on each of the other two. The remaining distance, (e)~22/57 to (g)~38/57, is facts that failed to carry across to new phrasing: a matching and admission problem.

\paragraph{Worked examples from the transfer-aware split.} The split is easiest
to see on concrete questions. The unit of relatedness is the phrase-level fact,
keyed by \obj{semantic\_family\_key}-style normalisation, and a held-out
question is admissible only if every fact it needs has at least one teacher in
train.

\emph{One fact, many wordings.} Held-out question~340, \emph{``Which are the
cards that have incredibly powerful foils,''} needs the single fact
\obj{powerful\_foils} (the predicate \obj{cardKingdomFoilId IS NOT NULL AND
cardKingdomId IS NOT NULL}). Its teachers in train are questions 394, 487
and 511:
\begin{itemize}\setlength\itemsep{1pt}
\item 394: \emph{``How many white bordered cards are powerful?''}
\item 487: \emph{``What is the percentage of incredibly powerful cards in the
set Coldsnap?''}
\item 511: \emph{``What is the most common visual frame effects among the
incredibly powerful foils?''}
\end{itemize}
The four questions share almost no SQL structure (a list, a count, a percentage, a mode) and three different phrasings of the same convention. What
they share is the fact, which is exactly what the mechanism transfers: a rule
whose name and description an LLM can match against new wording
(\S\ref{sec:split}).

\emph{Conjunctive coverage.} Held-out question~341, \emph{``What are the
borderless cards available without powerful foils?,''} needs \emph{two} facts:
\obj{powerful\_foils} (teachers as above) and \obj{borderless\_card} (teacher
370, \emph{``How many borderless cards are illustrated in Russian?''}). A
target enters the held-out set only when \emph{every} such fact has a teacher,
which is why the curve in \S\ref{sec:e3} can saturate: once the stream has
covered a target's last untaught fact, nothing further in train can help it.

\begin{table}[t]
\centering
\caption{\textbf{Database selection.} All 11 BIRD dev databases, annotated with phrase-level facts and screened on four requirements the protocol imposes, each computable before any learning runs: (C1) enough shared-fact structure to build a transfer split: splits were built for seven databases, and the four left unsplit have the fewest shared families at the lowest coverage (4--12 families, 5--35\% of questions), a judgment at the margin, since \obj{european\_football\_2} (12 families, 35\%) is comparable to \obj{formula\_1} (13, 33\%), which was split and then failed C4, and \obj{formula\_1}, the closest miss, scores 64/174 at day 0, in line with \obj{card\_games}, so it was not dropped for difficulty; (C2) at least 100 training questions, for a stream of five or more batches of 20; (C3) at least 40 held-out questions, so one answer moves the score by under 2.5 points; (C4) a live teacher population, at least half of train teaching some held-out fact. \emph{shared}: families carried by two or more questions; \emph{live}: families with a teacher in train and a target in held-out; \emph{ckpts}: stream checkpoints, as run for the three selected and as batches of 20 would give for the rest.}
\label{tab:screen}
\scriptsize
\setlength{\tabcolsep}{3.5pt}
\begin{tabular}{@{}lrrrrrrrrrl@{}}
\toprule
database & Q & families & shared & \% Q in shared fact & train\,/\,h.o. & teachers & live & pts\,/\,h.o.\ Q & ckpts & outcome \\
\midrule
\obj{card\_games} & 191 & 60 & 39 & 66.5 & 134\,/\,57 & 70 & 39 & 1.8 & 7 & \textbf{selected} \\
\obj{thrombosis\_prediction} & 163 & 142 & 78 & 90.2 & 118\,/\,45 & 102 & 71 & 2.2 & 6 & \textbf{selected} \\
\obj{toxicology} & 145 & 33 & 18 & 71.0 & 101\,/\,44 & 59 & 18 & 2.3 & 5 & \textbf{selected} \\
\obj{formula\_1} & 174 & 44 & 13 & 32.8 & 132\,/\,42 & 15 & 13 & 2.4 & 7 & fails C4 \\
\obj{financial} & 106 & 36 & 24 & 76.4 & 74\,/\,32 & 49 & 24 & 3.1 & 4 & fails C2, C3 \\
\obj{california\_schools} & 89 & 64 & 24 & 64.0 & 64\,/\,25 & 32 & 22 & 4.0 & 4 & fails C2, C3 \\
\obj{debit\_card\_specializing} & 64 & 18 & 12 & 82.8 & 45\,/\,19 & 34 & 12 & 5.3 & 3 & fails C2, C3 \\
\obj{european\_football\_2} & 129 & 29 & 12 & 34.9 & -- & -- & -- & -- & -- & fails C1 \\
\obj{student\_club} & 158 & 15 & 9 & 19.0 & -- & -- & -- & -- & -- & fails C1 \\
\obj{codebase\_community} & 186 & 20 & 11 & 17.7 & -- & -- & -- & -- & -- & fails C1 \\
\obj{superhero} & 129 & 17 & 4 & 5.4 & -- & -- & -- & -- & -- & fails C1 \\
\bottomrule
\end{tabular}
\end{table}

\emph{The same annotation on other warehouses.} The procedure is per-database and applies unchanged. On BIRD \obj{toxicology} it yields a 101/44 split over 33 families (18 transfer); e.g.\ held-out \emph{``How many triple type bonds are there?''} is one of seven targets of the fact \obj{triple\_bond} (the stored form \obj{bond\_type = '\#'}), taught by eight training questions, among them \emph{``Find the triple-bonded molecules which are carcinogenic''} and \emph{``Identify all connected atoms with a triple bond''}; one question (252) was placed in train rather than the target set for a quirk in its gold. On \obj{thrombosis\_prediction} it yields 118/45 over 142 families (71 transfer); e.g.\ held-out \emph{``For patients born between 1930 and 1940, how many percent of them were inpatient?''} is the single target of \obj{in\_patient} (the stored code \obj{Admission = '+'}), taught by eight admission-flag questions. Both splits carry the full protocol, reported alongside \obj{card\_games} throughout \S\ref{sec:results} (Tables~\ref{tab:base}--\ref{tab:e3}); the per-question annotations, family tables and split files for all three databases are released at \splitrepo.

\FloatBarrier
\end{document}